\documentclass[twoside]{article}

\usepackage[accepted]{aistats2021}
\usepackage{wrapfig,lipsum,booktabs}

\usepackage{setspace}
\usepackage{algorithm}
\usepackage{algorithmic}
\usepackage[utf8]{inputenc} 
\usepackage[T1]{fontenc}    
\usepackage{hyperref}       
\usepackage{url}            
\usepackage{booktabs}       
\usepackage{amsfonts}       
\usepackage{nicefrac}       
\usepackage{microtype}      
\usepackage{makecell}
\usepackage{tcolorbox}
\newcommand{\highlight}[1]{\colorbox{blue!10}{#1}}

\newcommand{\len}[1]{\mathit{len}(#1)}
\usepackage{amsmath}
\usepackage{todonotes}

\begin{document}

%

%

\twocolumn[

\aistatstitle{Neural Function Modules with Sparse Arguments:\\A Dynamic Approach to\\Integrating Information across Layers}

\aistatsauthor{Alex Lamb \And Anirudh Goyal  \And  Agnieszka S\l{}owik }
\aistatsaddress{Universite de Montreal\And Universite de Montreal \And University of Cambridge}

\aistatsauthor{Michael Mozer \And Philippe Beaudoin \And Yoshua Bengio }
\aistatsaddress{Google Research / University of Colorado \And Element AI \And Mila} 

\runningtitle{Neural Function Modules with Sparse Arguments}
\runningauthor{Lamb, Goyal, S\l{}owik, Mozer, Beaudoin, Bengio}
]
\begin{abstract}
Feed-forward neural networks consist of a sequence of layers, in which each layer performs some processing on the information from the previous layer. A downside to this approach is that each layer (or module, as multiple modules can operate in parallel) is tasked with processing the entire hidden state, rather than a particular part of the state which is most relevant for that module. Methods which only operate on a small number of input variables are an essential part of most programming languages, and they allow for improved modularity and code re-usability.  Our proposed method, Neural Function Modules (NFM), aims to introduce the same structural capability into deep learning. Most of the work in the context of feed-forward networks combining top-down and bottom-up feedback is limited to classification problems. The key contribution of our work is to combine attention, sparsity, top-down and bottom-up feedback, in a flexible algorithm which, as we show, improves the results in standard classification, out-of-domain generalization, generative modeling, and learning representations in the context of reinforcement learning.  
\end{abstract}

\section{Introduction}

\begin{figure*}[ht!]
    \centering
    \includegraphics[width=0.8\linewidth]{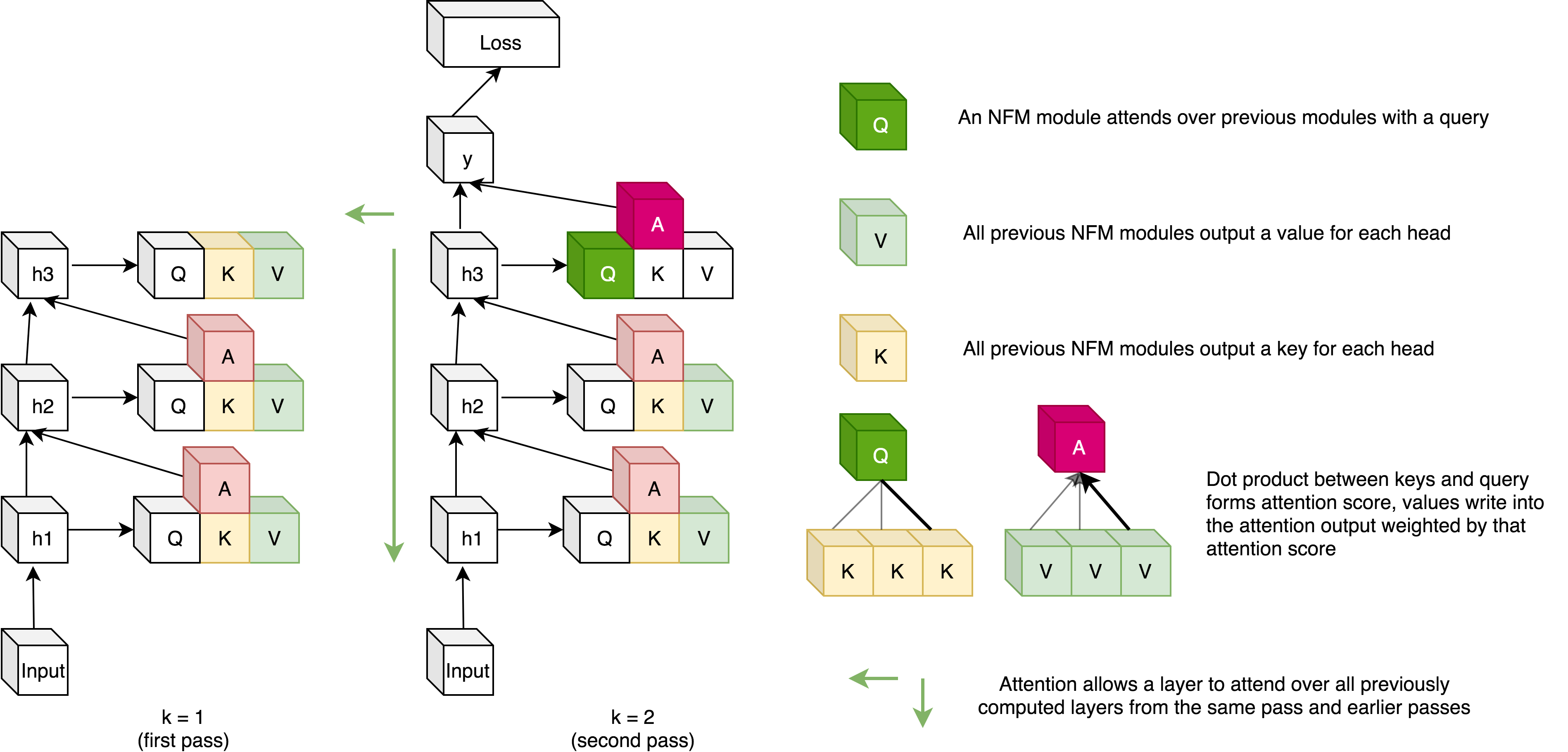}
    \caption{An illustration of NFM where a network is ran over the input twice ($\mathcal{K}=2$).  A layer where NFM is applied sparsely attends over the set of previously computed layers, allowing better specialization as well as top-down feedback.  }
    \label{fig:my_label}
\end{figure*}

Much of the progress in deep learning architectures has come from improving the ability of layers to have a specific focus and specialization.  One of the central drivers of practical progress in deep learning has been finding ways to make networks deeper and to create inductive bias towards specialization.  Indeed, the general trend in state-of-the-art techniques has been from networks with just a few layers to networks with hundreds or thousands of layers, where each layer has a much more specific role.  Iterative inference and generation \cite{marino2018iterative, jastrzkebski2017residual, greff2019multi} approaches also share this motivation: since they only require each pass to make a small change to an underlying representation.  This has been applied in iterative inference for generative models \cite{wu2019logan}.  

Perhaps the best example of this is residual networks (ResNets) \cite{He_2016_CVPR}, in which an additive skip-connection is placed between layers.  This allows a layer to use any other layer via the linear additive skip connections, yet this is a rigid and inflexible communication between layers.  Despite that limitation, residual networks have now become ubiquitous, and are now used in almost all networks where computational resources allow for a large number of layers.  


Other ideas in feed-forward networks have explored how to make individual layers more narrowly focused.  In the DenseNet \cite{huang2017densely}, each layer takes as input all previous layers within the same block, concatenated together.  This reduces the need to store redundant information in multiple layers, as a layer can directly use information from any prior layer given as input, even if it does not directly precede that layer.  Neural ODEs \cite{NIPS2018_7892} explored a continuous variant of ResNets, in which it can be directly shown that making each layer perform an even more narrowly focused computation (in this sense, meaning a smaller update) can improve accuracy.  



While these techniques have greatly improved the generalization performance of deep networks, in large part by making individual layers more specialized, we argue that they are still limited in that they take the entire current program state as input, rather than dynamically selecting input arguments.  To address this limitation, we take inspiration from computer programs, which are much richer and more flexible than today's deep networks. 
Computer programs are typically organized into methods, which perform a specific task, and only operate over a specific set of inputs which are relevant for that task.  This has several advantages over writing code without these well-contained methods.  One reason is that it leads to better separation of concerns and modularity.  A method which is written once can be reused in different programs, even if the overall purpose of the program is very different.  Another benefit to using methods is that it allows the programmer to avoid accidentally using or changing variables unrelated to the function. In this way a program could use a larger number of variables and have a greater amount of complexity, with a contained risk of introducing bugs.  Another advantage is that mistakes can be isolated down to a specific method within the program, which can lead to a more concentrated and efficient credit assignment.  

Additionally, methods in computer programs have the property that they are able to take arguments from the entire program state which is in scope.  Thus they may use either recently computed variables, or variables computed much earlier in the program.  This has some connection to bottom-up and top-down feedback signals from cognitive psychology, with recently computed variables or input streams being analogous to bottom-up signals and higher-level goals or descriptive variables being analogous to top-down signals.  This idea has seen some work in the deep learning community \cite{zamir2017feedback}, although many deep architectures (such as feed-forward networks) rely exclusively on bottom-up processing.

Our proposal is to allow modules (corresponding to computational blocks which can be executed in parallel or in sequence) in a deep network to attend over the previously computed modules from the network to construct their input.  This proposal, which we call \textit{Neural Function Modules (NFM)}, is motivated by analogy with functions in programming languages - which typically only take a few variables as arguments, rather than acting upon the entire global state of the program.  Likewise an NFM module selects the hidden states from the previously computed modules as its inputs, and may use its entire hidden state to focus on these few inputs.  The modules which come later in the network may attend to either that module or the modules preceding it.  

Additionally, to allow the NFM to consider both top-down and bottom-up signal specialization, we use a multi-pass setup, in which we run the networks forward pass multiple times, and allow the NFM to attend over all of the previously seen passes.  In a classifier, the later parts of the network from earlier passes correspond to ``top-down'' feedback, since they come from a network which has seen and has a compressed representation of the entire example.  Whereas attending to parts of the network closer to the input correspond to ``bottom-up'' feedback.  In general we consider a two-pass setup, but also perform analysis on single-pass and three-pass setups.  

We believe that the largest impact from adding this structure to deep learning will be in tasks where the question of when a layer is used is dynamic, which can occur as a result of the data distribution shifting during training, or the data distribution systematically differing between training and evaluation.  For example, the former occurs organically while training a GAN generator, since the objective from the discriminator is constantly evolving as the adversarial game progresses.  

Our newly proposed NFM module captures concepts of iterative inference, sparse dynamic arguments, and flexible computation, yet is straightforward to integrate into a variety of existing architectures.  We demonstrate this by integrating NFM into convolutional classifiers, GAN generators, and VAE encoders.  In each case, the structure of the NFM module is kept the same, but the details of the integration differ slightly.  

Our method offers the following contributions: 

\begin{itemize}
    \item Allowing layers to focus their entire output space size without limiting the information accessible to later modules.
    \item Making modules more specialized, by allowing them to focus on particular relevant inputs, rather than the entire hidden state of the network.  
    \item Providing a mechanism to allow for dynamic combination of top-down and bottom-up feedback signals. 
\end{itemize}

The key contribution of our work is to combine attention, sparsity, top-down and bottom-up feedback, in a flexible algorithm which can improve the results in the standard classification, out-of-domain generalization, generative modeling and reinforcement learning (as demonstrated in Section~\ref{experiments}
). Most of the work in the context of feed-forward networks combining top-down and bottom-up feedback is limited to classification problems. To the best of our knowledge, no work has combined these ingredients together in a unified architecture, that can be used to show improvements in various different problems, i.e classification, generative modelling, out of distribution generalization and learning representations in the context of RL.  

Our experiments show that these NFM modules can dynamically select relevant modules as inputs, leading to improved specialization of modules.  As a result, we show that this leads to improved generalization to changing task distributions.  We also demonstrate its flexibility, by showing that it improves performance in classification, relational reasoning, and GANs.

\begin{table*}[]
\centering
\caption{Desiderata and Related Work: showing how our motivations relate to prior work.}
\begin{tabular}{|l|l|}
\hline
Desiderata & Related Methods \\ \hline
Sparse Arguments & SAB \cite{ke2018sparse}, Sparse Transformer \cite{child2019generating} \\ \hline
Iterative Inference & LOGAN \cite{wu2019logan},GibbsNet \cite{lamb2017gibbsnet} \\ \hline
Dynamically Skipping Layers & SkipNet \cite{wang2018skipnet}  \\ \hline
Combining Top-Down and Bottom-Up Feedback & Deep Boltzmann Machines \cite{salakhutdinov2009deep} \\ \hline
\end{tabular}
\end{table*}

\section{Related Work}

\textbf{DenseNet:} These use concatenation to combine all of the layers within a block as inputs.  The largest difference between NFM and DenseNet is that NFM uses sparse attention, whereas DenseNet uses concatenation.  Another key difference is that NFM uses attention over many previously seen modules, even if they are of different sizes, whereas DenseNet only uses the handful of previously computed layers of the same spatial dimension. Most of the papers that have used DenseNets, have tested it on mostly computer vision problems (like classification), whereas NFM is a generic module that can be used for improving systematic generalization in relational reasoning, for classification as well as for generative modelling. An attentive variant of DenseNets was also proposed \cite{DBLP:journals/corr/abs-1805-11360}.

\textbf{Conditional Computation and Modularity:} Most of the  current deep learning systerms are built in the form of one big network, consisting of a layered but otherwise monolithic structure, which can lead to poor adaptation and generalization \cite{andreas2016neural, santoro2017simple, bahdanau2018systematic, bengio2019meta, goyal2019recurrent}. In order to address this problem, there have been attempts in modularizing the network such that a neural network is composed dynamically from several neural modules, where each module is meant to perform a distinct function \cite{andreas2016neural, shazeer2017outrageously, rosenbaum2017routing, goyal2019recurrent}.  The motivation behind methods that use conditional computation is to dynamically  activate only a portion of the entire network for each example \cite{bengio2013deep, wu2018blockdrop, fernando2017pathnet, mcgill2017deciding}. Recently \cite{hu2018gather, woo2018cbam, wang2018skipnet, chen2018gaternet,  veit2018convolutional} have proposed to use a learned gating mechanism (either using reinforcement learning or evolutionary methods) to dynamically determine when to skip in ResNet in a context dependent manner to reduce computation cost. In this line of work, no multiple modules are explicitly defined. Instead, the whole network is dynamically configured  by selectively activating model components such as hidden units and
different layers for each input example. Most of these works have been applied to computer vision problems such as image classification or segmentation or object detection. Our work is more related to such methods that dynamically decide where to route information but instead of skipping a particular layer or a unit, NFM dynamically decide where to \textit{query} information from (i.e., which layers to attend to), using sparse attention. Another key difference in the proposed method is that layers can attend to both the \textit{bottom-up}, as well as \textit{top-down} information which has been shown to improve systematic generalization as also evident in our experiments. Also, NFM is a generic module that can be used for other problems whereas most of these methods have been studied exclusively for image classification.

\textbf{Transformers:} The Transformer architecture uses attention over positions, but only on the previous layer.  NFM attends over layers, making it complementary with Transformers, yet the methods share a related motivation of allowing parts of the network to dynamically select their inputs using attention.  Future work can also investigate on integrating NFM module with transformers.

\textbf{Sparse Attention in RNNs: } Sparse Attentive Backtracking (SAB) \cite{ke2018sparse}  used sparse attention for assigning credit to a sparse subset of time steps in the context of RNNs leading to efficient credit assignment as well as efficient transfer. More recently, Recurrent Independent Mechanisms (RIMs) \cite{goyal2019recurrent} also use sparse attention for dynamically selecting a sparse subset of modules in an input dependent manner. Both SAB and RIMs uses sparse-attention in the context of RNNs, where the proposed method uses attention to dynamically integrate bottom-up as well as top-down information. 




\section{Neural Function Modules}

Our goal is to introduce the idea of Neural Function Modules (NFM) as a new way of composing layers in deep learning.  To that end, we start with desiderata motivating our design without going into architectural details.  Next, we give a detailed algorithmic description using a specific set of tools (such as top-k softmax \cite{ke2018sparse} as a way of implementing sparse attention), while we note that our concept is more general than this specific architecture.  We then describe how NFM can be integrated into various architectures.  



\subsection{Desiderata}




We lay out desiderata motivating our design, in the hope of defining a
class of architectures that share these goals:

\begin{itemize}
    \item Creating deep networks in which the communication between layers is dynamic and state-dependent, allowing it to be invariant to prior layers which are not relevant for it;  
    \item To allow a deep neural network to selectively route information flow around some layers, to break the bottleneck of sequential processing;
    \item To allow a model to dynamically combine bottom-up and top-down information processing using attention; and  
    \item To introduce more flexibility into deep architectures, by breaking the constraint that each layer needs to be a suitable input for the following layer.  For example, a layer which destroys fine-grained spatial information may be difficult to use earlier in the network, but could be valuable for later processing in deeper layers.  
\end{itemize}


\subsection{Proposed Implementation}

We describe our particular implementation of the Neural Functional Module concept which uses sparse attention, although we view the basic idea as being more general, and potentially implementable with different tools.  In particular, we introduce a new NFM module which is encapsulated and has its own parameters.  It stores every previously seen layer (from the forward pass) in its internal state, and uses the current layer's state as a query for attending over those previously seen layers.  This is described in detail at in Algorithm~\ref{alg:nfm_model}.  

\begin{algorithm}[]
    \caption{Neural Functional Module (NFM)}
   \label{alg:nfm_model}
   \begin{algorithmic}[1]
   \footnotesize
   \STATE {\bfseries Input:} An input $x$.  A number of passes $\mathcal{K}$.  A neural network with $N$ modules for all $k \in \{1 ... \mathcal{K}\}$: $f^{(1)}_{\theta_k}$, $f^{(2)}_{\theta_k}$, $f^{(3)}_{\theta_k}$, ...,$f^{(N)}_{\theta_k}$  
   \STATE $\mathcal{M} := \mathit{EmptyList}$
   \FOR{$k = 1$ to $\mathcal{K}$}
   \STATE $h^{(0)} := x$
   \FOR{$i = 1$ to $N$}
   \STATE $\tilde{\mathcal{M}} := \mathit{EmptyList}$
   \FOR{$j = 1$ to $i$}
   \STATE $\mathit{\tilde{\mathcal{M}}_j}.append(\mathit{rescale}(\mathcal{M}_j), \mathit{scale}(h^{(i-1)}))$
   \ENDFOR{}
   \STATE $R = \mathit{Attention}^{(i)}_{\theta_k}(\mathit{K}=\tilde{\mathcal{M}}, \mathit{V}=\tilde{\mathcal{M}}, \mathit{Q}=h^{(i-1)})$
   \STATE $h^{(i)} := f^{(i)}_{\theta_k}(\mathit{residual} = h^{(i-1)}, \mathit{input}=R)$
   \STATE $\mathcal{M}$.append($h^{(i)}$)
   \ENDFOR{}
   \ENDFOR{}
   \item[]
   
\end{algorithmic}

\end{algorithm}

\subsubsection{Mutli-Head Attention Mechanism}

We aimed to use the multi-head attention mechanisms from Transformers with as few modifications as possible.  We use linear operations (separate per module) to compute the key, value, and query for each module.  We then use softmax top-k attention \cite{ke2018sparse, goyal2019recurrent}, in which the attention only attends over the elements with the top-k highest attention scores.  Then we use one linear layer, followed by an activation, followed by a linear layer to project back to the original size.  Depending on the problem setting, we may or may not use batch normalization.

We now lay out the structure of the multi-headed attention mechanism which is used in Algorithm~\ref{alg:nfm_model}.  The $\mathit{Attention}^{(i)}_{\theta_K}$ function takes a set of keys and values to be attended-over: $K$, $V$, a query $q$, and a residual connection $R$.  The keys and queries are given dimension $d_k$ and the values are given dimension $d_v$.  It is specific to the layer index $i$ and the parameters for attention for the layer $i$: $(W_q, W_k, W_v, W_{o_1}, W_{o_2}, \gamma) = \theta^{(A)}_{i}$.  These $W$ refer to weight matrices and $\gamma$ is a learned scalar which is initialized to zero at the start of training.  Additionally a $\emph{k}$ to specify the k for top-k attention (typically k is set to a value less than ten).  An activation function $\sigma$ must also be selected (in our case, we used ReLu).  We place a batch normalization module directly before applying the activation $\sigma$.  

We set $\hat{Q} = qW_q$, $\hat{K} = KW_k$, $\hat{V} = VW_v$.  Then we compute $A = \mathit{Softmax}(\frac{\hat{Q}\hat{K}^T}{d_k})V$.  Afterwards the output hidden state is computed as $h_1 = \sigma(AW_{o_1})W_{o_2}$.  Then the final output from the attention blocked is multiplied by a $\gamma$ scalar and added to the residual skip-connection: $h_2 = R + \gamma h_1$.  The weighting with a $\gamma$ scalar was used in self-attention GANs \cite{zhang2018self}.  

\paragraph{Default Computation}
Additionally, we append a zero-vector to the beginning of the list of keys and values for the layers to be attended over.  Thus, a querying position on some (or all) of its heads may elect to look at these zeros rather than reading from an input element.  This is analogous to performing a function with a number of arguments which is less than the number of heads. 

\subsubsection{Rescaling Layers (Automatic Type Conversion)}
\label{sec:rescale}

When we query from a module of spatial dimension $\len{h_i}$, we consider attending over modules which may have either the same or different spatial dimensions $\len{h_j}$.  For simplicity, we discuss 1D layers, but the rescaling approach naturally generalizes to states with 2D or 3D structure.  

If $\len{h_i} = \len{h_j}$, then the two modules are of the same scale, and no rescaling is performed.  If $\len{h_i} > \len{h_j}$, then $h_j$ needs to be upsampled to the size of $h_i$, which we do by using nearest-neighbor upsampling.  The interesting case is when $\len{h_i} < \len{h_j}$, which is when downsampling is required.  The simplest solution would be to use a nearest-neighbor downsampling, but this would involve arbitrarily picking points from $h_j$ in the downsampling, which could discard interesting information.  Instead, we found it performed slightly better to use a SpaceToDepth operation to treat all of the points in the local window of size $\frac{\len{h_j}}{\len{h_i}}$ as separate positions for the attention.  Thus when attending over a higher resolution module, NFM can pick specific positions to attend over.

\subsection{Integrating NFM}

We have presented an encapsulated NFM module which is flexible and could potentially be integrated into many different types of networks.  Notably, in all of these integrations, the internal structure of the NFM module itself remains the same, demonstrating the simplicity and flexibility of the approach.  Additionally, using NFM introduces no additional loss terms.  





\section{Experiments}
\label{experiments}

\begin{table}
\centering
\caption{Classification Results (\% Test Accuracy) with NFM show consistent improvements across tasks and architectures.  All results use Input Mixup with $\alpha=1.0$ and the experimental procedure in \cite{verma2018manifold}}
\label{tab:classifier}
\scalebox{0.84}{
\begin{tabular}{lccc}
\hline
Methods & Base Arch. & Baseline &  NFM \\ \hline
CIFAR-10 & PreResNet18 & 96.27 $\pm$ 0.2  & \highlight{96.56 $\pm$ 0.09}   \\
CIFAR-10 & PreResNet34 & 96.79 $\pm$ 0.1 & \highlight{97.05 $\pm$ 0.1} \\
CIFAR-10 & PreResNet50 & 97.31 $\pm$ 0.1 & \highlight{97.58 $\pm$ 0.2} \\
CIFAR-100 & PreResNet18 & 78.15 $\pm$ 0.1 & \highlight{78.66 $\pm$ 0.3} \\
CIFAR-100 & PreResNet34 & 80.13 $\pm$ 0.3 & \highlight{80.77 $\pm$ 0.1} \\
Tiny-Imagenet & PreResNet18 & 57.12 $\pm$ 0.3& \highlight{58.32 $\pm$ 0.2}   \\   Imagenet & PreResNet18  & 76.72  & \highlight{77.1} \\
\hline

\end{tabular}
}
\end{table}

\begin{table}
\centering
\caption{Results on Atari games with NFM show improved scores with NFM used in the game-image encoder.}
\vspace{2mm}
\scalebox{0.85}{
\begin{tabular}{lcccc}
\hline
Game & Architecture & Baseline &  NFM \\ \hline
Ms. Pacman & ResNet18 & \highlight{6432}  &  6300 $\pm$ 3384  \\
Alien & ResNet18 & 12500 & \highlight{14382 $\pm$ 239}	\\   
Amidar & ResNet18 & 3923 & 	\highlight{3948 $\pm$ 23}	\\   
Q-bert & ResNet18 & 27433  & \highlight{33253 $\pm$ 2302}	\\   
Crazy Climber & ResNet18 & 333242 &	\highlight{334323 $\pm$ 2389}\\  
\hline

\end{tabular}
}
\label{tab:atari}
\end{table}



%

Our experiments have the following goals:  
\begin{itemize}
    \item Demonstrate that Neural Function Modules can improve results on a wide array of challenging benchmark tasks, with the goal of demonstrating the practical utility and breadth of the technique.
    \item To show that NFM addresses the bottleneck problem in the size of the hidden layers, by achieving drastically improved performance when the model is made narrow, with very few hidden units per module. 
    \item To show that NFM improves generalization when the train and test set differ systematically, as a result of improved specialization of the modules over sparse functional sub-tasks. 
\end{itemize}
See Appendix~\ref{sec:hyperparameters} for more details on experimental setup and hyperparameters.  





\subsection{GANs}
Intuitively, generating images with structured objects involves various sparse functional operations - for example creating a part such that it is consistent with another part, or generating a part with particular properties.  Based on this intuition we integrated NFM into  InfoMax GAN \cite{lee2020infomaxgan}, and we modify only the generator of the GAN by integrating NFM (the discriminator and all of the losses are kept the same).  In our integration, we use two passes $\mathcal{K}=2$ and we place an NFM module before each residual block.  Thus a total of 8 NFM modules are integrated.  We used $d_k=32$, $d_v=32$, and 4 heads for the attention.  For both CIFAR-10 and Imagenet our base generator architecture is a ResNet18.  For more details, see Appendix~\ref{sec:gan}.

We used the \cite{lee2020mimicry} GAN code base with default hyperparameters for all of our GAN experiments.  We only changed the architecture of generator to incorporate NFM.  Thus it might be the case that we could have achieved even better performance if we re-tuned the hyperparameters specifically for our the NFM architecture, yet in practice we found solid improvements even without doing this.  A significant improvement on GANs using NFM are shown in Table~\ref{tab:gan} as measured by Frechet Inception Distance (FID) \cite{heusel2017gans} and Inception Score (IS) \cite{salimans1606improved}. We integrate NFM directly into the Pytorch Mimicry codebase which contains a variety of techniques: SNGAN, SSGAN, InfoMax-GAN, and WGAN-GP. The NFM model outperforms all of these strong baselines (Table~\ref{tab:gan}).


\begin{table*}[h!]
\centering
\caption{Improved generation with GANs (no use of class labels) on CIFAR-10 and Tiny-Imagenet, outperforming many strong baselines on Inception Score (IS) and Frechet Inception Distance (FID).  We compare our NFM(InfoMax-GAN) against three external baselines: SNGAN \cite{miyato2018spectral}, SSGAN \cite{chen2019self}, and InfoMax-GAN  \cite{lee2020infomaxgan}.  }
\begin{tabular}{lcccc}
\hline
Methods & CIFAR-10 FID & CIFAR IS & Tiny-Imagenet FID & Tiny-Imagenet IS \\ \hline
SNGAN  & 16.77 $\pm$ 0.04 & 7.97 $\pm$ 0.06	& 23.04 $\pm$ 0.06 & 8.97 $\pm$ 0.12 \\
SSGAN & 14.65 $\pm$ 0.04 & 8.17 $\pm$ 0.06 	& 21.79 $\pm$ 0.09 & 9.11 $\pm$ 0.12 \\   
InfoMax-GAN & 15.12 $\pm$ 0.10 & 8.08 $\pm$ 0.08 & 20.68 $\pm$ 0.02 & 9.04 $\pm$ 0.10	\\    
NFM(InfoMax-GAN) & \highlight{13.15 $\pm$ 0.06} & \highlight{8.34 $\pm$ 0.02} & \highlight{18.23 $\pm$ 0.08} & \highlight{9.12 $\pm$ 0.09}	\\     
\hline

\end{tabular}
\label{tab:gan}
\end{table*}



\subsection{Stacked MNIST Generalization}
\label{mnist_gen}

We consider a simple multi-object classification task in which we change the number of object between training and evaluation, and verify how well the model is able to generalize.  Our reasoning is that if the model learns sparse functional modules for recognizing the objects, then it should be able to handle novel numbers of objects.  To keep this task as simple as possible, we construct synthetic 64x64 images containing multiple MNIST digits, and we train a convnet with the output as a multi-label binary classifier for each digit.  

For the integration with NFM, we used two passes ($\mathcal{K}=2$) and use a top-k sparsity of $k=5$.  Thus the NFM module is used 8 times with $d_k=16$, $d_v=16$, and 4 heads.  For more details and experimental setup, see Appendix~\ref{sec:stacked_mnist}).  


\begin{table}[h!]
\centering
\caption{Recognizing images with multiple mnist digits: training on one or three digits (top), one or five digits (bottom), provides evidence of improved specialization over the digit recognition sub-task (test accuracy \%).  }
\scalebox{0.95}{
\begin{tabular}{lcc}
\hline
Methods & Baseline & NFM \\ \hline
Trained (1,3) digits\\\hline
One Digit    &  $99.27 \pm 0.05$ & $99.23 \pm 0.03$      \\
Three Digits & $87.83 \pm 0.02$ & $87.78 \pm 0.46$ \\
\hline
Two Digits & $76.7 \pm 1.34$ & \highlight{$88.36 \pm 0.85$}\\
Four Digits  & $62.57 \pm 0.04$ & \highlight{$66.12 \pm 1.86$}  \\
Five Digits  & $24.27 \pm 1.02$  & \highlight{$29.48 \pm 3.12$}  \\ \hline
Trained (1,5) digits\\\hline
One Digit & $99.22 \pm 0.09$ & $99.16 \pm 0.04$       \\
Five Digits  & $68.87 \pm 4.25$  & $71.76 \pm 2.79$ \\
\hline
Two Digits & $74.69 \pm 2.88$ & \highlight{$84.01 \pm 6.06$} \\
Three Digits & $57.11 \pm 3.57$ & \highlight{$66.58 \pm 5.24$} \\
Four Digits  & $75.90 \pm 2.48$ & \highlight{$79.04 \pm 2.37$}  \\
\hline

\end{tabular}}
\label{tab:mnist}
\end{table} 

\subsection{Relational Reasoning}



In relational reasoning a model is tasked with recognizing properties of objects and answering questions about their relations: for example, ``is the red ball in front of the blue ball''.  This problem has a clear sparse functional structure which requires first recognizing the properties of objects from images and then analyzing specific relations, and thus we sought to investigate if NFM could improve results in this domain.  We used the Sort-of-CLEVR \cite{NIPS2017_7082} task to evaluate NFM in the context of visual reasoning and compositional generalization. We analyzed the effect of extending a generic convolutional baseline (CNN\_MLP) with NFM (specifically, NFM-ConvNet; Table \ref{tab:sortofclevr}). The baseline implementations and dataset generation follow description in the paper which introduced Sort-of-CLEVR \cite{NIPS2017_7082}. Similarly as in Section \ref{mnist_gen}, we use two passes ($\mathcal{K}=2$) and a top-k sparsity of $k=5$. We report mean test accuracy and standard deviation (over three trials) in Table \ref{tab:sortofclevr} and Table \ref{tab:sortofclevrsysgen}.



\begin{table}
\centering
\caption{Test accuracy on Relational reasoning (Sort-of-CLEVR) from images.}
\scalebox{0.9}{
\begin{tabular}{lcc}
\hline
Task & CNN & NFM(CNN) \\ \hline
Relational qst &  $68.26 \pm 0.61$ & \highlight{$74.95 \pm 1.58$}  \\
Non-relational qst & $71.78 \pm 9.3$ & \highlight{$79.37 \pm 2$} \\
\hline

\end{tabular}
}
\label{tab:sortofclevr}
\end{table}



Images in Sort-of-CLEVR consist of 6 randomly placed geometrical shapes of 6 possible colors and 2 possible shapes. There are 10 relational and 10 non-relational questions per image. Non-relational questions have two possible answers (random guess accuracy is 50\%). This also applies to relational questions of type 1 and type 2 (reasoning over distances between the objects). Relational questions of type 3 (count) have 6 possible answers. Therefore on average a random baseline for relational questions has accuracy of $\approx39\%$. While Sort-of-CLEVR is a simple dataset in its original form, it can be made substantially more difficult by introducing a distribution shift (Table \ref{tab:sortofclevrsysgen}). Table \ref{tab:sortofclevrsysgen} shows the results of omitting $N$ color-shape combinations from the training set and testing on the original $N=12$ combinations. We include the results of a strong baseline, Relation Networks (RNs) \cite{NIPS2017_7082}. Note that RNs contain a module specifically tailored for answering relational questions in this task.

\begin{table*}
\centering

\caption{Compositional generalization (Sort-of-CLEVR) to unseen variations, suggesting better specialization over uncovering attributes and understanding relations.  }
\vspace{1mm}
\scalebox{1.0}{
\begin{tabular}{lccc}
\hline
Number of hold-out combinations & CNN & Relation networks & NFM(CNN) \\ \hline

$N=1$: Relational qst & \highlight{$57.67 \pm 0.47$} & $50.67 \pm 0.94$ & $54.67 \pm  1.7$\\
$N=1$: Non-relational qst & $57 \pm 1.63$ & $48 \pm 0.82$ & \highlight{$59 \pm 0.81$} \\
$N=2$: Relational qst & \highlight{$47 \pm 3.27$} & $45 \pm 3.56$ & $46 \pm 1.41$ \\
$N=2$: Non-relational qst & $54.33 \pm 0.47$ & $43.33 \pm 7.31$ & \highlight{$57 \pm 1.63$} \\
$N=3$: Relational qst & $20.66 \pm 3$ &  \highlight{$40.67 \pm 1.7$} & $38.33 \pm 3.3$ \\
$N=3$: Non-relational qst & $42.33 \pm 1.25$ & $46 \pm 3.56$ & \highlight{$49.67 \pm 3.3$} \\
Average & $46.99$ & $45.61$ & \highlight{$50.78$} \\
\hline

\end{tabular}
}
\label{tab:sortofclevrsysgen}
\end{table*} 

The accuracy of both baselines decreases towards random performance with a distribution shift between training and test data. Our model outperforms the simple baseline and RNs in the non-relational set of questions, suggesting that NFM improves the stage of recognizing object properties even in the presence of a distribution shift. While generalization to out-of-distribution samples remains challenging when combined with relational reasoning, NFM might alleviate the need for additional fully connected layers such as those used in RNs. For more details regarding the setup we ask the reader to refer to Appendix~\ref{sec:relational_reasoning}.









\subsection{Classification and Generalization to Occlusions}

We evaluated NFM on the widely studied classification benchmarks CIFAR-10, CIFAR-100, Tiny-Imagenet, and Imagenet and found improvements for all of them, with the goal of demonstrating the utility and versatility of NFM (Table~\ref{tab:classifier}).  We used a two-pass setup with $k=5$, and we trained both the NFM and baseline models with Mixup \cite{zhang2017mixup}.  We integrated with base architectures of PreActResNet18, PreActResNet34, and PreActResNet50 \cite{he2016identity}.  

As an ablation study on the classifier, we tried training with NFM normally, but at test time changed the attention values to be random (drawn from a Gaussian).  We found that this dramatically hurt results on CIFAR-10 classification (96.5\% to 88.5\% test accuracy).  This is evidence that the performance of NFM is dependent on the attention selectively picking modules as inputs.  

In addition to improving classification results, we found that classifiers trained with NFM had better robustness to occlusions (not seen at all during training).  Evidence from neuroscience shows that feedback from frontal (higher) brain areas to V4 (lower brain areas) is critical in interpreting occluded stimuli. \cite{fyall2017dynamic}.  Additionally research has shown that neural nets with top-down feedback better handle occluded and cluttered displays. \cite{spoerer2017recurrent}.  Using our normally trained NFM model on CIFAR-10 with PreActResNet18, we improve test accuracy with occlusion boxes of size 16x16, with a single occlusion box per image, from 82.46\% (baseline) to 84.11\% (NFM).  Our occlusion used the same parameters as from the Cutout paper \cite{devries2017improved}. For more details refer to Appendix~\ref{sec:classification_task}.

\subsection{Atari}

The Atari 2600 game environment involves learning to play simple 2D games which often contain small modules with clear and sparsely defined behavior patterns.  For this reason we sought to investigate if using an encoder with NFM modules could lead to improved results.  All compared methods used the same ResNet 
backbone, input preprocessing, and an action repeat of 4. Our NFM integration used two passes ($\mathcal{K}=2$) and top-k sparsity of $k=4$.  We chose games that require some degree of planning and exploration as opposed to purely reactive ones: Ms. Pacman, Frostbite, Alien, Amidar, Hero, Q-bert, Crazy Climber. We choose this set of games, as it was previously used by \cite{vezhnevets2016strategic}. We integrate NFM into the resnet encoder of a Rainbow IQN \cite{dabney2018implicit,hessel2018rainbow}.  We use keysize $d_k$ of 32, value size $d_v$ of 32, 4 heads, two-passes $\mathcal{K}=2$, and top-k sparsity of $k=4$.  Thus the NFM module is added at four places in the integration.  After integrating NFM, we kept the same hyperparameters for training the IQN model.   We use exactly the same setup as in \cite{rltime} for RL algorithm, as well as other hyper-parameters not specific to the propose architecture. We show substantially improved results on four of these five games (see Table~\ref{tab:atari}
). 







\subsection{Analysis of Hyperparameters}

On CIFAR-100 classification (PreActResNet34) we jointly varied the keysize, valsize, and number of heads used for the NFM process (Table~\ref{tab:keyvalhead}).  

\begin{table}[h!]
\centering
\caption{Varying the key dimension, value dimension, top-k sparsity, and number of heads used for the attention process, when running PreActResNet34 on CIFAR-100 classification.  Note that all results outperform the baseline accuracy of $80.13\%$. }
\vspace{2mm}
\begin{tabular}{lccccc}
\hline
Heads & Top-k & $d_k$ & $d_v$ & Test Accuracy (\%) \\ \hline
2 & 3 & 8 & 16 & 80.37 \\
2 & 3 & 8 & 32 & 80.47 \\
2 & 3 & 8 & 64 & 80.41 \\
2 & 4 & 8 & 16 & 80.26 \\
2 & 4 & 8 & 32 & \highlight{80.77} \\
2 & 4 & 8 & 64 & \highlight{80.52} \\
4 & 3 & 16 & 16 & 80.31 \\
4 & 3 & 16 & 32 & 80.26 \\
4 & 3 & 16 & 64 & 80.27 \\
4 & 3 & 32 & 32 & 80.40 \\
4 & 4 & 16 & 32 & \highlight{80.55} \\
4 & 4 & 32 & 64 & 80.33 \\
\hline
\end{tabular}
\label{tab:keyvalhead}
\end{table} 

On the relational reasoning task, we tried using one-pass $\mathcal{K}=1$, and we achieved test accuracy of $73.07 \pm 1.17$ on relational questions, which is better than the baseline's $68.26 \pm 0.61$ accuracy, but worse than the NFM with two-passes $\mathcal{K}=2$ (accuracy of $74.95 \pm 1.58$ on relational questions). We also saw an improvement thanks to introducing attention sparsity. In the reported results, both baseline CNN and NFM(CNN) use 24 initial channels. We also looked at the relation between the number of model parameters, test accuracy and adding/removing NFM in an image classification task (Appendix~\ref{sec:classification_task}).





\section{Conclusion}

The central concept behind the success of deep learning is that models should consist of multiple components (layers), each performing specific functions.  Many of the most successful ideas in deep learning have the purpose of allowing individual layers to serve a more specific and incremental role. However, we note that most neural networks still process all of the layers in a fixed sequence, giving each layer the previous layer as input.  We have instead proposed an alternative setup, which we called Neural Function Modules (NFM), which is inspired by functions in programming languages, which operate over specific arguments. Our main contribution lies in a new algorithm design, which connects several ideas that are important in deep learning (attention, sparsity, specialized modules, top-down and bottom-up feedback, long-range dependencies). The proposed implementation of these ideas (Neural Function Modules) is a generic and highly flexible architecture. While it is improved by NFM, increased standard test accuracy on classification tasks such as ImageNet was not the main focus of our work.  We have shown that the proposed method substantially improves the performance across many different tasks (including systematic generalization), and we have shown ways in which this opens up new opportunities for architecture design - by removing the constraint that each layer must serve as input for the successive layer.

\newpage

\clearpage
\bibliography{refs}

\appendix

\onecolumn

\section{Hyperparameter Analysis}
\label{sec:hyperparameters}

We performed some additional experiments to measure the importance of a few critical hyperparameters for NFM.  

We tried replacing the normal re-scaling process (Section~\ref{sec:rescale}) from NFM and replaced it with a simple nearest-neighbor re-scaling.  On CIFAR-10 PreActResNet18 classification, the average test accuracy dropped from $96.56\%$ to $96.47\%$




On the relational reasoning task, we tried using one-pass $\mathcal{K}=1$, and we achieved test accuracy of $73.07 \pm 1.17$ on relational questions, which is better than the baseline's $68.26 \pm 0.61$ accuracy, but worse than the NFM with two-passes $\mathcal{K}=2$ (accuracy of $74.95 \pm 1.58$ on relational questions). We also saw an improvement thanks to introducing attention sparsity. In the reported results, both baseline CNN and NFM(CNN) use 24 initial channels.

\section{Classification Task}
\label{sec:classification_task}

We considered classification on CIFAR-10, CIFAR-100, Tiny-Imagenet, and Imagenet.  We consider variants on the PreActResNet architecture \cite{he2016identity}.  For CIFAR-10, CIFAR-100, Tiny-Imagenet, and Imagenet and followed the same hyperparameters and configuration as the input mixup baseline in \cite{verma2018manifold}.  On all datasets except for Imagenet, we trained for 600 epochs, with a starting learning rate of 0.1, and dropped the learning rate by 10x at 200 epochs, 400 epochs, and 500 epochs.  We averaged our obtained test accuracy over the last 10 epochs.  We used input mixup with a rate of $\alpha=1.0$ \cite{zhang2017mixup}, except on Imagenet, where we used $alpha=0.5$.  For these experiments, we used a keysize of 32 and valsize of 32, with 4 heads.  We integrated NFM modules after each residual block.  Thus we used 8 NFM modules (4 in the first pass and 4 in the second pass).  We used a top-k sparsity for the attention of $k=5$. 

We also investigated if a single pass of NFM improves the accuracy regardless of the number of parameters in the model. We used the recent image classification task from \url{https://github.com/ElementAI/synbols}. Preliminary results are in Table~\ref{tab:synbols}.

\begin{table}[h!]
\centering

\caption{Default Synbols dataset (100k examples).}
\begin{tabular}{lccccc}
\hline
Num parameters & Num inital planes & NFM? & Test Accuracy (\%) \\ \hline
 206k & 12 & Yes & \highlight{83.14} \\
 177k & 12 & No & 81.89 \\
 757k & 24 & Yes & \highlight{84.03} \\
 698k & 24 & No & 82.61 \\
 5M & 64 & Yes & \highlight{87.63} \\
 4M & 64 & No & 85.2 \\
\end{tabular}
\label{tab:synbols}
\end{table} 

While more investigation is needed, in image classification adding NFM seems to scale better (in terms of test accuracy) than increasing the model size by adding more layers. Without NFM, the accuracy increases by $0.72\%$ at the cost of $698k - 177k = 521k$ additional parameters, whereas by adding NFM we get an increase of $1.25\%$ at the cost of adding $206k - 177k = 29k$ parameters. The effect persists for $12, 24, 64$ initial planes (Table~\ref{tab:synbols}).

\section{Stacked MNIST Generalization}
\label{sec:stacked_mnist}

For stacked mnist, we consider a base CNN with nine convolutional layers, with every other layer having a stride of two.  Each convolutional layer had a kernel size of 3.  The first convolutional layer had 32 channels, which we doubled every time the resolution is reduced, leading the final number of channels to be 512.  Each batch contains a certain number of digits per image, either (1 or 3) digits or (1 or 5) digits.  As a result of the batches having variable characteristics, we elected to remove the batch normalization layer when training on this task.  

We trained all models with Adam with a learning rate of 0.001 for 300 epochs, and report the test accuracy from the epoch with the highest validation accuracy.  

\begin{figure}
    \centering
    \includegraphics[width=0.6\linewidth]{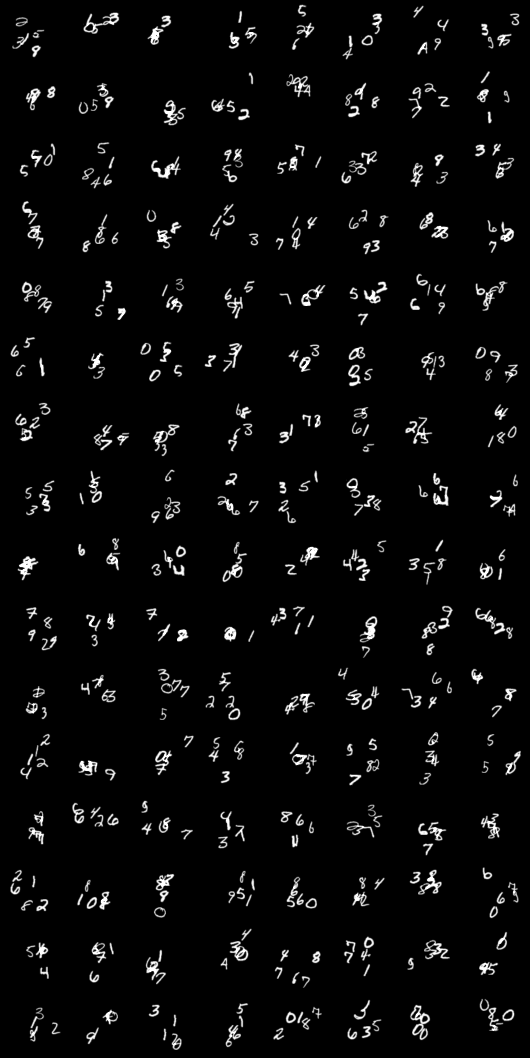}
    \caption{Examples of the stacked mnist digit dataset with 5 digits per image.  }
    \label{fig:mnist_ex}
\end{figure}

The images in our stacked MNIST task are 64x64, and some examples with 5 mnist digits are shown in Figure~\ref{fig:mnist_ex}.  Each digit is reduced to a size of 16x16 by nearest-neighbor downsampling and then pasted into a random position within the frame.  

\clearpage

\section{Relational Reasoning}
\label{sec:relational_reasoning}

\begin{figure}[h!]
    \centering
    \includegraphics[width=0.8\linewidth]{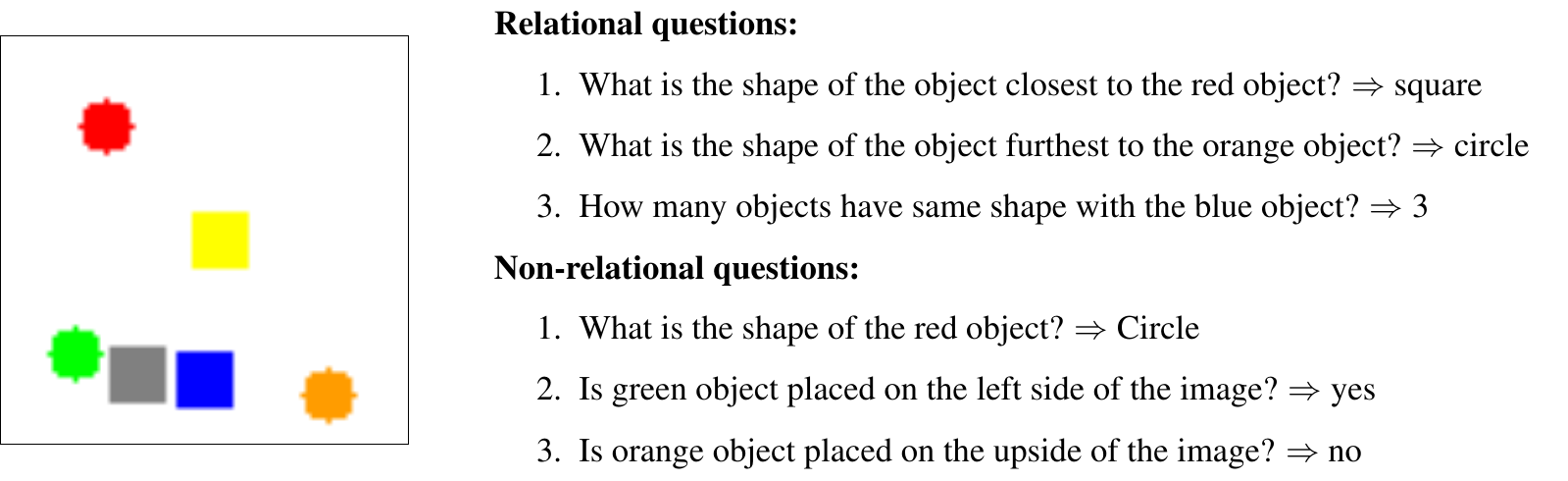}
    \caption{A sample from the Sort-of-CLEVR dataset.}
    \label{fig:soc_ex}
\end{figure}

Figure \ref{fig:soc_ex} shows a sample (image, question) from the Sort-of-CLEVR dataset. Each image is paired with 10 relational and 10 non-relational questions.  We use the exact CNN baseline architecture from \cite{antoniou2018dilated} along with the same experimental setup.  We train all models for 250 epochs with the Adam optimizer with a learning rate of 0.001.  

\section{Atari Reinforcement Learning}
\label{sec:atari}

We integrate NFM into the resnet encoder of a Rainbow IQN \cite{dabney2018implicit,hessel2018rainbow}.  We use keysize $d_k$ of 32, value size $d_v$ of 32, 4 heads, two-passes $\mathcal{K}=2$, and top-k sparsity of $k=4$.  Thus the NFM module is added at four places in the integration.  After integrating NFM, we kept the same hyperparameters for training the IQN model.   We use exactly the same setup as in \cite{rltime} for RL algorithm, as well as other hyper-parameters not specific to the propose architecture.


\section{Generative Adversarial networks}
\label{sec:gan}

We integrated NFM into an Infomax-GAN for both CIFAR-10 and Tiny-Imagenet.  We elected to integrate NFM into the generator only, since it is computationally much cheaper than using it in both the generator and the discriminator, as multiple discriminator updates are done for each generator update.  The original Infomax-GAN 32x32 generator consists of a linear layer from the original latents to a 4x4 spatial layer with 256 channels.  This is then followed by three residual blocks and then a final convolutional layer.  Our integration applies NFM after this first spatial layer and after the output of each residual block.  Since we use two passes $\mathcal{K} = 2$, the NFM module is applied a total of 8 times (4 in each pass).  

The only change we introduced was the integration of NFM and the two-pass generator.  Aside from that, the hyperparameters for training the GAN are unchanged from \cite{lee2020mimicry}.

\section{Computational Resources}

\textbf{Resources Used:} It takes about 4 days to train the proposed model on Atari RL benchmark task for 50M timesteps.


\end{document}